**D. V. Lande**
Institute for information recording, Kiev, Ukraine
dwl@visti.net

**Visualization of features of a series of measurements with one-dimensional cellular structure**

*This paper describes the method of visualization of periodic constituents and instability areas in series of measurements, being based on the algorithm of smoothing out and concept of one-dimensional cellular automata. A method can be used at the analysis of temporal series, related to the volumes of thematic publications in web-space.*

Визуализации особенностей рядов измерений посвящены многочисленные исследования. В частности, вейвлет- и дисперсионный анализ [1-3] позволяют выявлять гармонические составляющие, тренды, локальные особенности. Предлагаемый метод, основанный на алгоритме сглаживания пиковых значений и популярной концепции одномерных клеточных автоматов [4], позволяет выявлять стационарные области, области с резкими относительными (возможно небольшими) скачками значений, периодичности временного ряда. С помощью этого метода не детектируются абсолютные амплитудные всплески, однако он хорошо показал себя на «изрезанных» структурах данных, близких к фрактальным. К таким данным относятся, в частности, временные ряды, связанные с объемами публикаций в веб-пространстве по определенным темам.

Рассматривается следующая модель. Каждому значению ряда измерений $\xi(t)$ (обозначим, $X_0 = \xi(t)$) соответствует одна клетка клеточного автомата. По ряду измерений строится сглаженный по специальному алгоритму ряд $X_1$. Затем ряду $X_1$ ставится в соответствие ряд $X_2$ (получаемый из $X_1$ по тому же алгоритму сглаживания) и т.д. В соответствии с алгоритмом сглаживания пиков, значения, которые принимают элементы рядов измерений $x_k(t) \in X_k$ ($k$ – шаг сглаживания, $t$ – номер элемента ряда измерений) составляют:

$$x_k(t) = \begin{cases} x_{k-1}(t), & \text{if} \quad x_{k-1}(t) \leq \dfrac{x_{k-1}(t-1) + x_{k-1}(t+1)}{2}; \\ \dfrac{x_{k-1}(t-1) + x_{k-1}(t+1)}{2}, & \text{if} \quad x_{k-1}(t) > \dfrac{x_{k-1}(t-1) + x_{k-1}(t+1)}{2}. \end{cases}$$

Цвет клетки одномерной клеточной структуры, соответствующей $X_k$, белый, если $x_k(t)$ совпадает с $x_{k-1}(t)$, в противном случае – черный. Таким образом, каждой клетке соответствует значение $x_k(t)$ и значение цвета. (*Необходимо отметить, что такую* *систему нельзя считать каноническим клеточным автоматом, так как в общем случае клеткам может соответствовать бесконечное множество значений).*

Таким образом, алгоритм сглаживания пиков и визуализации можно представить в следующем виде:

Шаг 1. Рассчитываются значения сглаженного временного ряда в соответствии приведенной выше формулой.

Шаг 2. Рассчитываются значения цветов соответствующих клеток, которые отображаются.

Шаг 3. Шаг итерации увеличивается на единицу.

Шаг 4. Если заданное заранее количество шагов превышает текущий шаг итерации, то происходит переход к шагу 1.

Если область значений представляют собой выпуклое вверх множество, то визуальное представление клеточных автоматов принимает вид сплошной черной полосы (рис. 1: вертикальная ось – номер шага итерации, а горизонтальная ось – номер элемента временного ряда).

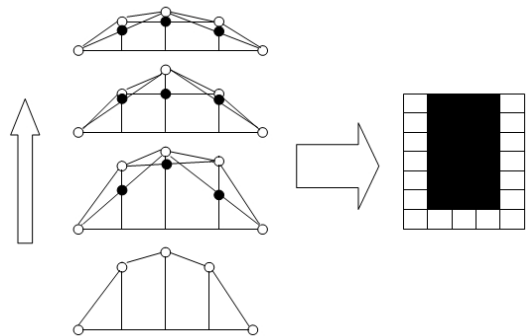

Рис. 1 – Выпуклое вверх множество точек

Единичные всплески значений в ряде измерений (рис. 2а) и области изрезанности (рис. 2б) могут вызывать появление структур типа «шахматной доски».

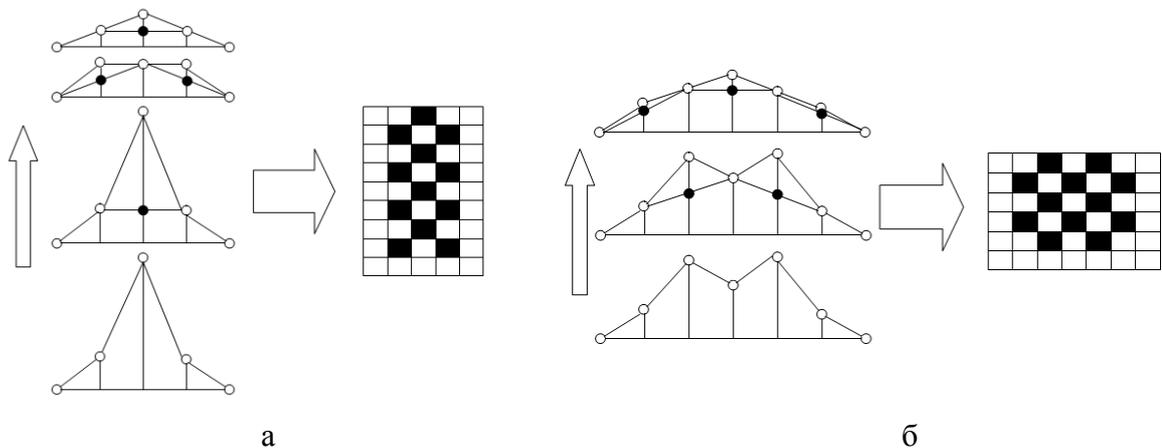

а                                                                   б

Рис. 2 – Появление структур типа «шахматной доски»

Диаграммы, формируемые в результате визуализации в соответствии с предложенным алгоритмом позволяют выявлять периодические составляющие, это можно продемонстрировать на примере двух функций $y = \sin(x)$ и $y = x\sin(x)$ (рис 3а и 3б, соответственно. Верхняя часть каждого из рисунков – исходные данные).

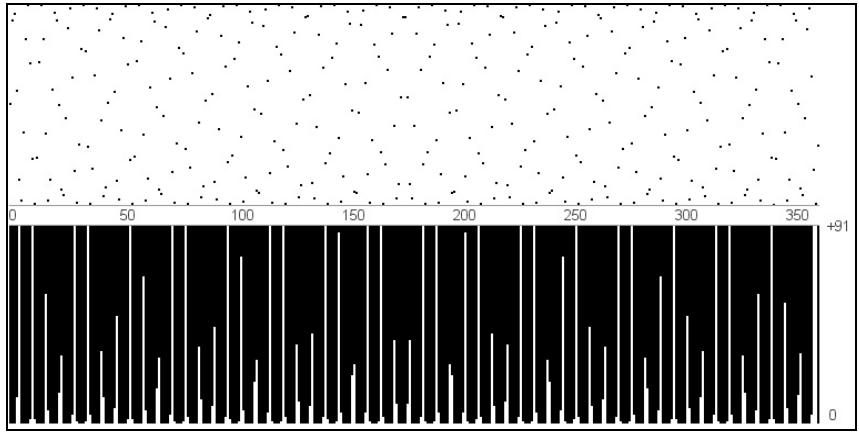

а

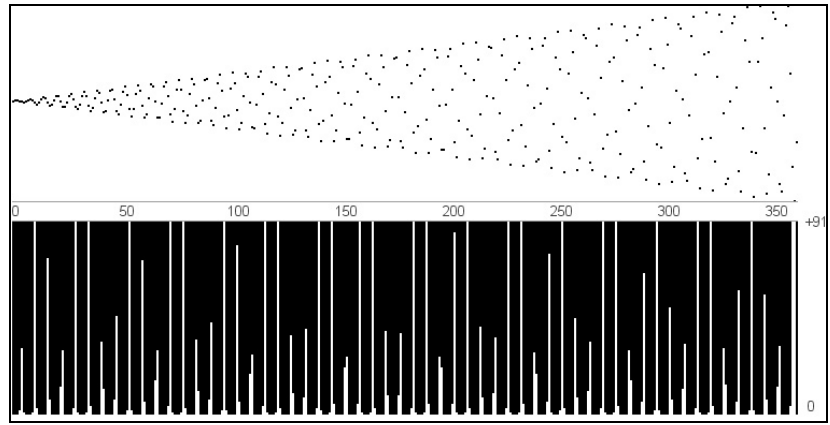

б

Рис. 3 – Отображение простых периодических составляющих

Функции, содержащие несколько гармонических составляющих, позволяют выявлять отдельные из них (пример: $y = \sin(x) + \cos(3x)$, рис 4).

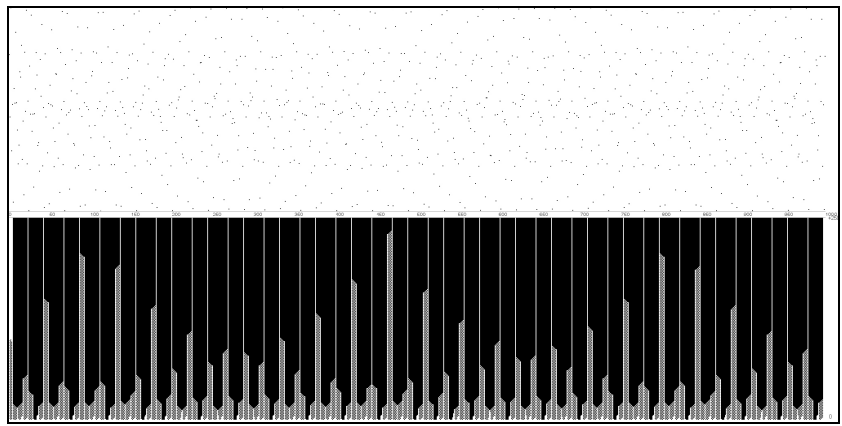

Рис. 4 – Отображение комплексных гармонических составляющих

Отображение реального временного ряда измерений, соответствующего посуточным объемам публикаций в веб-пространстве по некоторой заданной теме (точки ряда – объемы публикаций за сутки) представлено на рис. 5. Здесь четко отслеживаются недельные периодичности публикаций (минимумы – праздники, субботы и воскресенья) и периоды нестабильности.

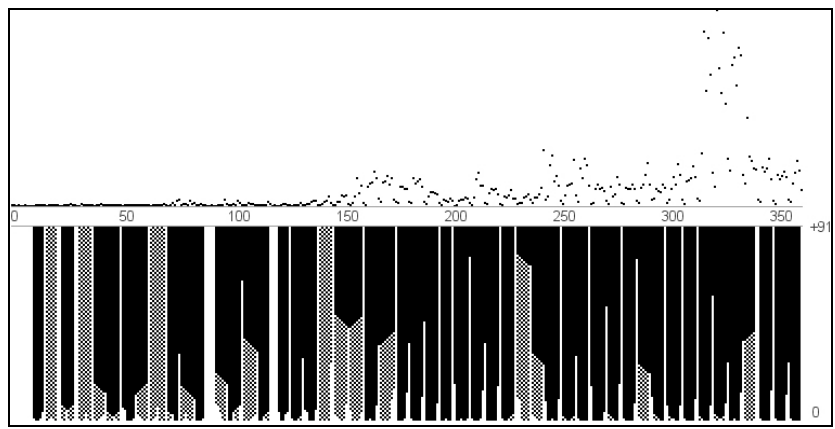

Рис. 5 – Отображение ряда динамики публикаций

Предложенный метод является относительно простым в программной реализации, так как базируется на алгоритме сглаживания пиков и концепции клеточных автоматов. Он позволяет визуально выявлять единичные и нерегулярные «всплески», резкие колебания значений количественных показателей в разные периоды времени. Метод испытывался при анализе временных рядов, связанных с объемами публикаций в веб-пространстве по определенным темам [5]. По-видимому, он может эффективно применяться при анализе временных рядов в таких областях как экономика и социология.